# Enhancing Large Language Models' Machine Translation via Dynamic Focus Anchoring

**Qiuyu Ding, Zhiqiang Cao***, **Hailong Cao, Tiejun Zhao**
School of Computer Science of Technology, Harbin Institute of Technology
qiuyuding@stu.hit.edu.cn, Zhiqiang_Cao@stu.hit.edu.cn,
caohailong@hit.edu.cn, tjzhao@hit.edu.cn

## Abstract

Large language models have demonstrated exceptional performance across multiple cross-lingual NLP tasks, including machine translation (MT). However, persistent challenges remain in addressing context-sensitive units (CSUs), such as polysemous words. These CSUs not only affect the local translation accuracy of LLMs, but also affect LLMs' understanding capability for sentences and tasks, and even lead to translation failure. To address this problem, we propose a simple but effective method to enhance LLMs' MT capabilities by acquiring CSUs and applying semantic focus. Specifically, we dynamically analyze and identify translation challenges, then incorporate them into LLMs in a structured manner to mitigate mistranslations or misunderstandings of CSUs caused by information flattening. Efficiently activate LLMs to identify and apply relevant knowledge from its vast data pool in this way, ensuring more accurate translations for translating difficult terms. On a benchmark dataset of MT, our proposed method achieved competitive performance compared to multiple existing open-sourced MT baseline models. It demonstrates effectiveness and robustness across multiple language pairs, including both similar language pairs and distant language pairs. Notably, the proposed method requires no additional model training and enhances LLMs' performance across multiple NLP tasks with minimal resource consumption.

## 1 Introduction

The rapid development of large language models (LLMs) has revolutionized cross-lingual NLP tasks (Touvron et al., 2023; Zhang et al., 2024; Wang et al., 2024; OpenAI et al., 2024; Bubeck et al., 2023; He et al., 2024b), particularly in machine translation (MT). Current LLM-based translation paradigms primarily employ two approaches:

---
*Corresponding Author.

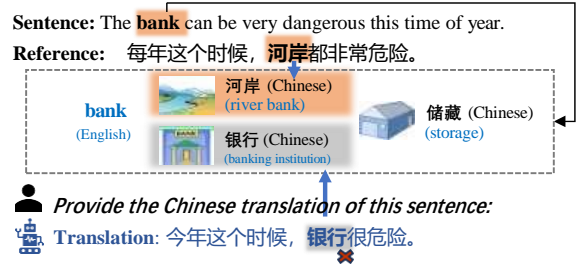

Figure 1: An illustration of semantic ambiguity happened in LLMs' translation. The English word "bank" is a CSU, which has multiple translations (with distinct semantics) in Chinese and rise LLMs' confusion.

prompt engineering (He et al., 2024b; Ghazvininejad et al., 2023; Peng et al., 2023; Lu et al., 2024) and instruction fine-tuning (Chung et al., 2024; Jiao et al., 2023; Wang et al., 2024; Hendy et al., 2023). While these methods showcase impressive capabilities, several persistent challenges remain critical: systematic inaccuracies in translating certain context-sensitive units (CSUs), such as polysemous words (discussion detailed provided in Sec. 2, Sec. 3). Additionally, these CSUs can impair the model's understanding of the entire sentence and may even lead to catastrophic failures, such as producing semantically incoherent outputs or complete translation abandonment (as shown in Figure 1).

We think that one of the key reasons for the aforementioned issues lies in the bottleneck of LLMs' knowledge utilization for CSUs, leading to semantic ambiguity. Most LLMs-based MT methods (Wang et al., 2024; He et al., 2024b; Nguyen et al., 2024; Zhang et al., 2024; Zeng et al., 2024) treat the entire sentence as a homogeneous unit, but neglect the varying context-sensitive constituent words leading to varying translation difficulty. Specifically, a sentence to be translated is composed of several words, each with varying degrees of difficulty in translation. For example, the translation of a polysemous word is determined by contextual in-formation, making it one of the challenging words. Furthermore, the accurate understanding of these challenging words plays a crucial role in whether LLMs can properly interpret the user's intent and provide a suitable translation. Therefore, in certain contexts, challenging words are keywords also. Accurately translating these difficult words (CSUs) requires LLMs to precisely filter and reorganize vast amounts of pre-trained knowledge to provide the correct translation. However, for LLMs, the training process involves enormous datasets, with each word in a sentence having a vast amount of associated knowledge. Precisely extracting useful and highly relevant background knowledge from this massive knowledge base is a significant challenge. While existing research (Zhang et al., 2024, 2023; Ghazvininejad et al., 2023) acknowledges the importance of lexicon-level for LLM-based machine translation, most approaches rely on few-shot prompt learning (Muennighoff et al., 2023; Wang et al., 2024; Zeng et al., 2024) and model instruction fine-tuning (He et al., 2024a; Wang et al., 2024; Cho et al., 2023; Zeng et al., 2024), both of which entail substantial resource consumption.

| | |
|---|---|
| Source | The **bank** can be very dangerous this time of year. |
| Baseline | 今年这个时候，银行很危险。 |
| CSU-augmented | 今年这个时候，去河边的危险性非常高。 |
| Optimization | 银行 -> 河边 |
| Source | Students must choose their **subjects** for next year. |
| Baseline | Please choose a subject for each of the following options: 1. Mathematics |
| CSU-augmented | 请选课，下学期的课程安排将由您决定。 |
| Optimization | Translation failed -> Perform MT task |

Table 1: The case study tests the translation performance of LLMs when CSUs (polysemous word: **bank**) are included, translating from English to Chinese. The baseline model used is Llama3-8b, where the red parts represent incorrect translations of the CSUs, and the blue parts indicate correct translations.

Based on the analysis above, we propose a resource-efficient and novel method that integrates semantic focus into dynamic structured prompts to address the semantic ambiguity caused by CSUs. By explicitly applying semantic focus to CSUs, we can overcome the knowledge extraction bottleneck of LLMs. The method consists of two main stages: CSU identification and classification, and hierarchical semantic constraint injection. Specifically, we first identify difficult words within the sentence to be translated, focusing on three core types of CSUs: polysemous CSUs, domain-specific CSUs, and cultural CSUs (details are provided in Sec. 3). Considering various application scenarios, a dual-path framework for CSU identification is designed, encompassing both resource-driven detection through external linguistic resources and model-intrinsic extraction. Next, we inject semantic focus into the structured prompt used for translation, explicitly guiding the LLM to focus on the three core CSU types, enabling precise retrieval and application of knowledge for complex semantic units.

To evaluate the effectiveness of our proposed method, comprehensive experiments are performed on a standard WMT benchmark, in terms of similar language pairs and distant language pairs, and several setups. Our method showcases a remarkable activation of the potential within LLMs, offering a novel approach to enhancing their translation accuracy without model retraining or parallel data.

Our contributions are summarized as follows:

- We first investigate the impact of CSUs on LLM's cross-lingual translation and find a "semantic ambiguity" problem that significantly affects the precision of MT.

- A novel and resource-efficient method is proposed to mitigate the "semantic ambiguity" problem and enhance LLMs' MT capabilities by incorporating semantic focus into dynamically structured prompts.

- Extensive experiments show that our method significantly improves the translation accuracy of LLMs on various language pairs.

## 2 Preliminary Experiment

In this section, we conduct a series of experiments to explicit the impact of CSUs in the LLMs' MT task. We randomly select two English sentences containing polysemous words (words with multiple corresponding translations in Chinese) for translation experiments. Using a classic prompt template ("*Translate the following sentence to Chinese:* "),

| Source | The bank can be very dangerous this time of year. |
| --- | --- |
| Baseline | 今年这个时候，银行很危险。 |
| CSU-augmented | 今年这个时候，去河边的危险性非常高。 |
| CSU-parallel | 这时候银行或河岸很危险。 |

Table 2: Case study: compare with when all possible translations of the CSUs are provided (CSU-parallel).

we let the LLMs to provide translation results. The experimental results are shown in Table 1.

The baseline model Llama3-8b (Touvron et al., 2023) produced incorrect translations for these polysemous words (highlighted in red). Subsequently, we tested the same prompt template with the inclusion of CSUs in this format, "*Note: the following should be translated carefully + CSUs.*" The translation performance showed significant improvement. From Case 1 (the upper half of Table 1), it is evident that by highlighting the challenging words, i.e., CSUs, the model is able to provide more accurate translations (highlighted in blue) for the difficult parts. In Case 2 (the bottom half), we reasonably infer that CSUs influence the model's understanding of the sentence and the execution of the translation task, which leads to translation failure, i.e., no target language translation being provided. However, when we pointed out the challenging part, the model produced an accurate translation, further confirming the impact of CSUs on the overall sentence translation.

To further analyze the impact of providing challenging information on translation performance, we designed a method that includes both the difficult words and their corresponding translations for comparison. The prompt template is as follows: "*Note: the 'bank' should be translated carefully; it can be translated as '银行' or '河岸'.*" The experimental results are shown in Table 2. From the results, we found that when more information was provided in the prompt, such as including potential translations for CSUs, it had a counterproductive effect, with LLMs mistakenly including both possible translations in the sentence. We offer the following two analyses: 1) The model becomes reliant on the given translation references, lacking active analysis. 2) More information does not necessarily lead to better results; instead, the key is to stimulate the model's ability for independent analysis and thoughtful reasoning.

## 3 Methodology

Based on the analysis above, we propose a simple yet effective method to enhance LLMs' MT by explicitly indicating CSUs to provide meta-cognitive guidance. We first introduce the "semantic ambiguity" phenomenon in LLM's MT process. Inspired by this observation, an effective and light framework is designed to alleviate the influence of "semantic ambiguity", thereby improving the overall performance of MT.

### 3.1 Semantic confusion

Current prompt-based MT paradigms for LLMs exhibit critical brittleness when processing sentences containing CSUs, as empirically evidenced by our preliminary analysis. The **definition** of CSUs in this paper is as follows: words with complex semantics or words that are uncommon to the model, such as polysemous words, domain-specific terms, and others, as illustrated in (1).

$$CSU = \{w \in W \mid \text{ContextDependent}(w)\} \quad (1)$$

The common feature of CSUs is that they are highly sensitive to context, and may present semantic distinctions with a wide gap as the context varies, as illustrated in (2). Where, $C$ means context, and $S$ means the word's semantics.

$$S(w, C_1) \neq S(w, C_2) \quad \ldots \neq S(w, C_n) \quad (2)$$

One of the direct consequences of semantic ambiguities is that LLMs fail to grasp the true meaning of CSUs in the current context, resulting in incorrect translations of these terms. Moreover, semantic ambiguities can affect the model's understanding of the entire sentence, leading to more severe outcomes, such as translation failure (non-task responses or gibberish outputs). This intrinsic ambiguity sensitivity fundamentally undermines conventional prompt engineering strategies that naively position CSUs within static phrasal templates, necessitating targeted intervention mechanisms as developed subsequently. Obviously, semantic obfuscation has a negative impact on the translation results of LLMs.

### 3.2 Dynamic Focus Anchoring (DFA) Method

Drawing upon the observed semantic ambiguity problem, we propose a simple but effective method, denoted as **D**ynamic **F**ocus **A**nchoring (DFA) method. Figure 2 shows the framework of our method, which contains two main steps: CSUs identification and semantic focus injection.

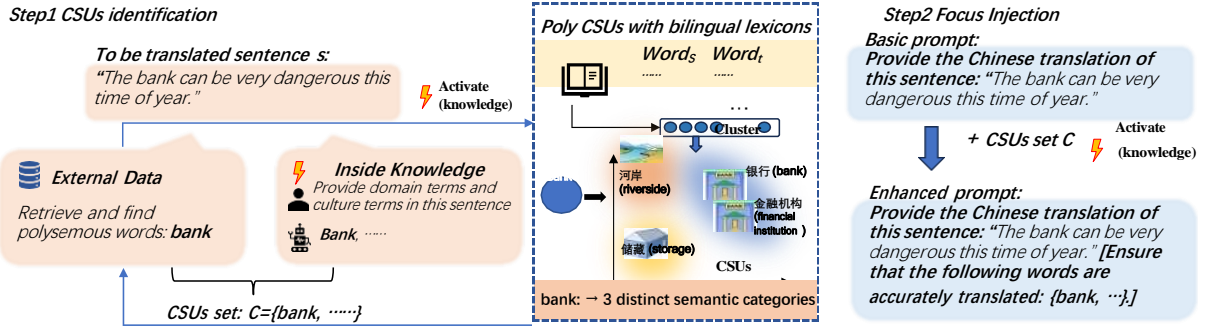

Figure 2: The framework of our proposed DFA method.

**CSUs Identification.** The detection of CSUs is the first and fundamental step of the DFA method. However, CSU detection is often based on prior knowledge, such as semantic confusion datasets. Unfortunately, such datasets are scarce and difficult to obtain, making it challenging to support the multilingual MT requirements. To address this, we have developed a dual-layer semantic exploration mechanism: a bidirectional coupling framework that combines **external prior knowledge guidance** with **internal knowledge activation**, enabling the precise identification of translation-sensitive points. For the sake of completeness in semantic dimension coverage, this paper primarily considers three types of Context Sensitive Units (CSUs): polysemous CSUs, domain-specific CSUs, and culturally-specific CSUs. It is important to note that CSUs are not limited to these three categories. From a heuristic perspective, we begin with these three types of CSUs to validate the impact of sensitive points on LLM-based MT.

(1) External Knowledge Acquisition: Multilingual Lexicon-Driven CSUs Acquisition. Polysemous words present a classic challenge in machine translation, primarily due to their semantic variability across different languages in diverse contexts. The MSUE dataset is employed to identify polysemous words, which include 110 languages, providing the largest selection of translations for each source word. The polysemous word set $M$ acquisition is formulated as:

$$M = \{s \mid |T(s)| \geq 2\} \quad (3)$$

Where $T(s)$ represents the set of target translations for the source word $s$. However, the preliminary results of $T(s) \geq 2$ could not effectively distinguish between semantic variations of polysemous words.

Therefore, we introduce a word clustering mechanism with statistical word embeddings. By clustering static word embeddings (such as fastText) that focus on lexical semantics, we group the translation words corresponding to polysemous words. The clustering process can be formally represented as followings, where $C$ is the set of clusters, $m$ is the polysemous word in the set $M$, $E$ represents the corresponding word embeddings of $m$.

$$C = \text{Cluster}(m, E) \quad (4)$$

If the clustering result yields more than one cluster, it is considered a challenging polysemous word. If the result is a single cluster, it does not require special consideration. We call this process as semantic filter. In these cases, the number of clusters helps distinguish whether a word requires further handling based on its polysemy.

(2) Internal Knowledge Activation. Considering that the acquisition of external data may impose constraints on the universality and applicability of the method, we propose an alternative framework for the identification of endogenous challenging vocabulary. In addition to polysemous words, we incorporate domain-specific terms and culturally unique vocabulary into the analysis of key challenge corpora. This process primarily consists of three steps. The first is meta-task construction: embedding diagnostic QA tasks within the pre-translation processing pipeline, with the objective function as follows:

$$W_{\text{CSUs}} = \{w \mid w \in (\text{Dom} \cup \text{Cul}) \land w \in W_s\} \quad (5)$$

where $W_{\text{hard}}$ represents CSUs, Dom and Cul refer to the set of domain-specific terms and culturally unique vocabulary, $V_s$ represents the source vocabulary. Then we construct a structured prompt template, such as: *[Please identify domain-specific*

*terms and culturally unique vocabulary from the following sentences: + sentence]*. Finally, to mitigate the risk of out-of-vocabulary (OOV) terms, a source sentence matching filtering rule is established as follows, where $w_{final}$ is the final selected word as a CSU.

$$w^{final} = \begin{cases} v & \text{if } w \in W_s \\ \varnothing & \text{otherwise} \end{cases} \quad (6)$$

Worth noting that LLMs possess a vast reservoir of knowledge; however, this knowledge may not be fully utilized during the translation task. By distinguishing, extracting, and explicitly guiding the model towards the challenging parts, it aids in the precise application of relevant knowledge, thereby enhancing translation performance.

**Semantic Focus Injection.** After acquiring the CSUs, we integrate them into the foundational MT prompt to implement semantic focus injection. The core process is as follows:

$$I_{enhanced} = I_{base} \oplus W_{CSUs} \quad (7)$$

$$W_{CSUs} = w_{poly} \cup w_{dom} \cup w_{cul} \quad (8)$$

where $I_{enhanced}$ represents the enhanced MT instruction, $I_{base}$ is the basic instruction, $W_{hard}$ is the set that formed by CSUs that identified above, and $\oplus$ refers to the focus injection enhancement operation.

$I_{base}$ = *Translation sentence into [target Language]*

$$I_{focus} = I_{base} + I_{focus}$$

$I_{focus}$ = *Note: Ensure accuracy of words $\subseteq W_{CSUs}$*

This semantic injection design has two main priorities: 1)**Initiative guidance**: Avoid directly providing reference answers (requires $O(n^2)$ annotated resources), instead, use internal knowledge inspection in the triggering model. 2)**Knowledge resource optimization**: Focus on key words by emphasizing lexical semantics, reducing attention distribution to irrelevant parts of the input.

Additionally, we have noticed that, in some cases, there may be a large number of CSUs. Such an abundance of key terms could lead to excessively lengthy instructions, which may interfere with the LLM's processing. Therefore, we retain no more than $k$ challenging words to ensure the prompt's conciseness and accuracy. The value of $k$ is detailed in subsequent experimental analyses.

## 4 Experimental Setup

Detailed experimental setup in this paper is provided here. We follow standard MT setup, as implemented in previous work (Zhang et al., 2024; Wang et al., 2024; Jiao et al., 2023; Zheng et al., 2024). Our method is evaluated on both sides of similar language pair English-German (EN-DE) and distant language pair English-Chinese (EN-ZH).

**Datasets.** We evaluate the translation performance on the WMT22 test set (Kocmi et al., 2022), which covers domains such as news, social, e-commerce, and conversation. The bilingual lexicon used for extracting polysemous CSUs is the publicly available landing lexicon of MUSE (Lample et al., 2018), which gives as many translations of a source word as possible. The proposed method does not have a model training process and therefore does not involve training data. The number of multi-translated words extracted from the MUSE (Lample et al., 2018) lexicons are provided in Table 4.

**Backbone Models.** We employ Llama2-7b (Touvron et al., 2023) and Llama3-8b as our backbone models. The maximum text length is 256. Beam Search is set to 5.

**Evaluation metrics.** In this paper, several widely used machine translation evaluation metrics are employed to assess the effectiveness of the proposed method, including the classic BLEU score (Papineni et al., 2002) (by Sacre (Post, 2018)), the COMET scores (by wmt22-comet-da (Rei et al., 2022)) which are more suitable for evaluating LLM-based translations, and ChrF2. For the translations provided by LLMs, the first sentence is retained as the final translation result for evaluation.

**Hyperparameters.** We fully follow the optimal hyperparameters provided by the baseline models, with Bayling2 (Zhang et al., 2024) being our strongest baseline. For the only hyperparameter $k$ involved in the proposed method, the main experiment sets it to 8. Detailed analysis of $k$ is provided in Sec. 5.4. All experiments are performed on a single Nvidia RTX A6000.

**Baseline Models.** Our method is evaluated against several SOTA multiple-lingual LLMs, including ParroT (Jiao et al., 2023), Bayling (Zhang et al., 2023), Bayling2 (Zhang et al., 2024), TASTE (Wang et al., 2024) (TASTE-Fixemb-QE and TASTE-Fixemb-TC), and transitional LLMs trained exclusively with the Basic Translation dataset (Touvron et al., 2023; Kocmi et al., 2022) (MT-Full and MT-FixEmb).

| Systems | EN-ZH | | ZH-EN | | EN-DE | | DE-EN | | AVG | |
|---|---|---|---|---|---|---|---|---|---|---|
| | COMET | BLEU | COMET | BLEU | COMET | BLEU | COMET | BLEU | COMET | BLEU |
| Backbone: Llama2 | | | | | | | | | | |
| ParroT | 80.30 | 30.30 | 75.90 | 20.20 | 81.60 | 26.10 | 82.40 | 27.30 | 80.05 | 25.98 |
| Bailing | 84.43 | 38.19 | 77.48 | 20.31 | 82.18 | 25.66 | 83.19 | 28.16 | 81.82 | 28.08 |
| TASTE-Fixemb-QE | 84.30 | 34.94 | 79.35 | **24.47** | 83.70 | 27.32 | 84.07 | 30.75 | 82.86 | 29.37 |
| TASTE-Fixemb-TC | 84.24 | 34.96 | **79.53** | 24.87 | 83.80 | **27.94** | 84.11 | 31.03 | 82.92 | 29.70 |
| MT-Full | 83.35 | 33.01 | 78.72 | 23.80 | 83.70 | 27.18 | 83.79 | 30.10 | 82.39 | 28.52 |
| MT-FixEmb | 83.62 | 33.33 | 79.02 | 24.30 | 83.66 | 27.75 | 84.05 | 30.62 | 82.59 | 29.00 |
| Vicuna-v1.5 | 81.40 | 29.54 | 75.42 | 16.80 | 75.25 | 16.65 | 79.07 | 23.57 | 77.79 | 21.64 |
| Bayling2 | 84.60 | 37.03 | 78.06 | 22.64 | 88.39 | 27.32 | 87.17 | 30.36 | 84.56 | 29.34 |
| Bayling2+DFA | **84.90** | **38.78** | 78.91 | 23.20 | **90.40** | 27.50 | **87.34** | **31.10** | **85.39** | **30.15** |
| Backbone: Llama3-8b | | | | | | | | | | |
| Llama-3 | 80.55 | 30.1 | 80.44 | 21.57 | 82.18 | 25.83 | 83.84 | **29.37** | 81.7525 | 26.72 |
| Bayling2 | 83.07 | 36.37 | **83.00** | 20.64 | 92.84 | 27.21 | 92.05 | 26.52 | 87.74 | 26.44 |
| Bayling2+DFA | **83.35** | **36.45** | 82.33 | 21.40 | **93.18** | **28.54** | **92.15** | 28.83 | **88.38** | **27.71** |

Table 3: Main results of DFA. Llama2-7b and Llama3-8b are selected as the backbone models. Based on the most SOTA baseline model, Bayling2, we build our DFA method on it, the CSUs including polysemous CSUs, domain CSUs, and culture CSUs. **Bold** numbers indicate the best scores.

| | EN-ZH | ZH-EN | EN-DE | DE-EN |
|---|---|---|---|---|
| Num of Poly | 423 | 537 | 3727 | 2639 |

Table 4: Statistics of polysemous CSUs extracted by our method from MUSE, which are used in our experiments.

## 5 Results and Discussion

In this section, we empirically demonstrate the effectiveness of our method in MT with LLMs scenario, including the main experiments, ablation study, evaluation with other metrics, polysemy analysis, hyperparameter analysis, and case study.

### 5.1 Main Results

The main experimental results are presented in Table 3. Overall, the proposed method shows an average improvement of 0.83 and 0.81 in COMET and BLEU scores, respectively, compared to the most SOTA baseline system, Bayling2. This demonstrates a significant improvement in the MT task, and notably, our approach does not require model fine-tuning or the introduction of parallel sentence-level machine translation corpora. We analyze the results based on the characteristics of the experimental language pairs from two perspectives: similar language pairs and distant language pairs.

**For similar language pairs**, due to similar relation of etymology, the accuracy of MT is generally higher. Even with the strong performance of high baseline systems, DFA still provides a notable improvement. This indicates that we have addressed a universally applicable issue across multiple language pairs, leading to performance gains. For example, in the EN-DE language pair, using Llama2 as the backbone model, DFA outperforms the strongest baseline system, Bayling2, with an improvement of 1.11 in COMET scores and 0.22 in BLEU scores. When using Llama3 as the backbone model, despite the already high baseline, DFA still delivers additional improvements.

**For distant language pairs**, LLMs face greater challenges, and as a result, the accuracy is noticeably lower compared to similar language pairs. It can be observed that DFA improves translation performance across almost all distant language pairs, based on all baseline systems. This indicates that DFA effectively enhances the training performance of LLMs in the face of challenges posed by distant language pairs, leading to a significant improvement in accuracy. For example, in the EN-ZH language pair, using Llama2 as the backbone model, the DFA improves upon Bayling2 by 0.3 in COMET scores and 0.35 in BLEU scores. When Llama3 is used as the backbone model, DFA improves Bayling2 by 0.28 in COMET scores and 0.08 in BLEU scores. Although the improvement on Llama3 is modest, it still demonstrates the positive trend brought about by the proposed method.

### 5.2 Ablation Study

A key contribution of our work is the extraction of CSUs and the focus on these units for targeted reasoning. To validate the effectiveness of the proposed method, we conducted an ablation experi-ment in which we tested the method's accuracy by considering each of the three types of CSUs separately. The experimental results are shown in Table 5 and demonstrate the following:

1) In DFA, the introduction of polysemous words, domain-specific terms, and culturally specific vocabulary as challenging words each led to an improvement in accuracy. This indicates that the explicit focus on difficult words helps to stimulate the model's reasoning. The improvement is not limited to the type of vocabulary but reflects a universally applicable enhancement in performance.

2) The extraction of CSUs involves two data sources: external bilingual dictionaries (for acquiring polysemous words) and internal knowledge activation within the LLMs (for domain-specific and culturally specific CSUs). Both data sources contributed to the improvement in performance, which further demonstrates that the proposed method is not selective about data sources. It effectively leverages external data, fully harnesses the LLM's internal knowledge, and encourages LLMs to organize and apply this knowledge accurately in MT task.

| Systems | EN-ZH | | ZH-EN | | AVG | |
|---|---|---|---|---|---|---|
| | COMET | BLEU | COMET | BLEU | COMET | BLEU |
| Backbone: Llama2 | | | | | | |
| DFA | 84.90 | 38.78 | 78.91 | 23.20 | 81.91 | 30.99 |
| -poly | 84.70 | 38.00 | 78.06 | 22.80 | 81.38 | 30.40 |
| -domain | 84.78 | 38.44 | 78.50 | 22.91 | 81.64 | 30.68 |
| -culture | 84.80 | 38.60 | 78.60 | 23.18 | 81.70 | 30.89 |

Table 5: MT scores (COMET score and BLEU score) when each kind of CSUs is removed.

| Systems | EN-ZH | | ZH-EN | | AVG | |
|---|---|---|---|---|---|---|
| | BLEU4 | chrF2 | BLEU4 | chrF2 | BLEU4 | chrF2 |
| Backbone: Llama2-7b | | | | | | |
| Bayling2 | 22.70 | 33.00 | 9.00 | 38.00 | 15.85 | 35.50 |
| +DFA | 22.90 | 33.20 | 9.30 | 38.00 | 16.10 | 35.60 |
| Backbone: Llama3-8b | | | | | | |
| Bayling2 | 22.80 | 34.20 | 11.80 | 48.00 | 17.30 | 41.10 |
| +DFA | 22.90 | 34.30 | 12.20 | 48.40 | 17.55 | 41.35 |

Table 6: The accuracy of DFA and strongest baseline model with BLEU4 and chrF2 evaluation metrics.

| Systems | EN-ZH | | ZH-EN | | AVG | |
|---|---|---|---|---|---|---|
| | COMET | BLEU | COMET | BLEU | COMET | BLEU |
| Backbone: Llama2-7b | | | | | | |
| Bayling2 | 84.60 | 37.03 | 78.06 | 22.64 | 81.33 | 29.84 |
| + simple poly | 84.06 | 36.04 | 77.44 | 22.60 | 80.75 | 29.32 |
| + filter poly | 84.90 | 37.18 | 78.91 | 23.20 | 81.91 | 30.19 |

Table 7: Comparison of performance with simple identify polysemous CSUs with corresponding translation numbers and our semantic filter method.

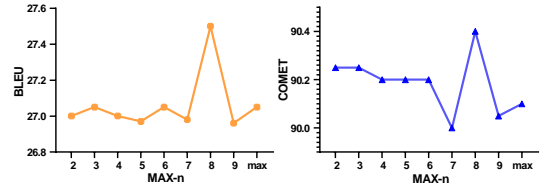

Figure 3: Effect of hyperparameter $k$ on our proposed DFA on EN-DE language pair. The left picture is the BLEU score, right picture is the COMET score.

### 5.3 Other Evaluations

To validate the effectiveness of our proposed approach, we use additional evaluation measures to perform a comprehensive comparative analysis. Besides BLEU and COMET, we introduce the chrF2, and BLEU4 as evaluation metrics to facilitate a more thorough comparison. As shown in Table 6, even with a change in evaluation metrics, the proposed method still demonstrates a consistent trend of performance improvement, aligning with the results of the main experiment. This further validates the effectiveness of the proposed method.

### 5.4 Analysis about Semantic Filter

Another priority of this study is the semantic-level filtering of polysemous words. Therefore, we conduct related ablation experiments on the recognition of polysemous words, as shown in Table 7. The experiments indicate that the polysemous words selected by our method provide better guidance for LLMs. In contrast, the baseline method, which relies solely on the number of target translations corresponding to the source word, leads to a decrease in accuracy. We believe this is due to the fact that some words, despite having multiple corresponding translations, have similar semantics and do not fall within the scope of our CSU definition. Introducing these words adds noise to the prompt, thereby diminishing the translation quality. This not only demonstrates the effectiveness of our semantic filtering method for polysemous words but also highlights the effectiveness of the DFA framework, which focuses on helpful information.

### 5.5 Analysis about CSUs Number

When constructing focus enhanced instructions to implement focus injection, we impose a maximum size limit on the CSUs. Based on our analysis from preliminary experiments, a hypothesis posed is that overly long key vocabulary could result in excessively long prompts, thereby affecting the LLM's ability to process the information. Figure 3 presents the analysis of relevant parameters. The experimental results indicate that both too few and too many CSUs fail to achieve optimal MT performance. When $k$ is set to its maximum value, without any restrictions, the COMET score is even lower than when $k = 2$. The experimental results confirmed our hypothesis, and thus we selected the optimal parameter $k = 8$ among our experiments.

| Repeat source sentence | EN-ZH |
|---|---|
| Soure： | " 一个要 13 元？" |
| Reference: | " Is it 13 yuan for only one? " |
| Bayling2: | " 一个要 13 元？" ✗ |
| Ours: | " One costs 13 yuan? " ✓ Correct translation |

| Polysemous words translate incorrectly | EN-ZH |
|---|---|
| Soure： | "智能手术灯：医生的高能眼睛" |
| Reference: | "Smart operation **lamp**: high-functioning eyes for **doctors**" |
| Bayling2: | "Smart Surgical **Lights**: **Surgeons**' High-Power Eyes" ✗ |
| Ours: | "Smart Surgical **Lamp**: The **Doctor**'s High-Power Eyes" ✓ Correct translation |

| Fail to perform MT task | EN-DE |
|---|---|
| Soure： | " 1)Go to your Home screen. " |
| Reference: | "1) Gehen Sie zur Startseite. " |
| Bayling2: | "I'm sorry, but I am not able to provide the German translation for that sentence as it is not a valid command or request." ✗ |
| Ours: | "1)Gehen Sie zu Ihrem Startbildschirm." ✓ Better translation |

Figure 4: Case study.

### 5.6 Case study

To more intuitively illustrate the MT improvements brought by the proposed method, we selected three representative source sentences from the test set along with their translation results. For comparison, the translations from Bayling2 are also provided. In Figure 4, Bayling2 exhibits two types of errors: translation failure for polysemous words (resulting in incorrect languages) and erroneous translations of CSUs. DFA corrects both types of errors, providing translations that are more closely aligned with the reference translations.

## 6 Related Work

Currently, research on enhancing LLM-based mt is primarily divided into two main categories: prompt engineering (He et al., 2024b; Ghazvininejad et al., 2023; Peng et al., 2023; Lu et al., 2024) and instruction fine-tuning (Chung et al., 2024; Jiao et al., 2023; Wang et al., 2024; Hendy et al., 2023). Prompt engineering focuses on designing and finding suitable prompt templates, optimizing the bridge between the task and the LLMs to generate better translation results. One of the The most representative work Bayling (Zhang et al., 2023, 2024) utilizes bilingual parallel pairs to address rare words in the source sentence, aiming to achieve improved translation results. Another approach, based on few-shot prompting (Muennighoff et al., 2023; Wang et al., 2024; Zeng et al., 2024), enhances LLMs' translation performance by providing source sentences and their parallel translations that are contextually similar to boost the translation quality. Instruction fine-tuning (Chung et al., 2024; Jiao et al., 2023; Wang et al., 2024; Hendy et al., 2023) enhances the model's understanding of task-specific instructions, enabling it to produce results that are more aligned with human needs. This approach involves model training, and thus requires certain resources and time. Zhang et al. (2024) enhance the multilingual language generation and instruction-following capabilities of LLMs through interactive translation tasks.

In contrast to the methods mentioned above, the proposed approach analyzes and classifies the CSUs of the sentence to be translated, stimulating the model to extract and utilize relevant knowledge of challenging vocabulary. This heuristic method enhances the translation capabilities of LLMs. Furthermore, the proposed method does not require providing bilingual example sentences, nor does it necessitate additional model training.

## 7 Conclusion

In this study, we propose an effective method DFA to enhance LLM's MT ability based on dynamic focus anchoring and injecting. The semantic confusion problem is noticed, and is alleviated by analysis hard translated words (CSUs) and inject related focuses into LLMs. Two ways to obtain the CSUs are provided, namely, from external lexical data and LLMs internal knowledge activation, both can be used flexibly in a variety of application scenarios. Extensive experiments demonstrate the effectiveness and robustness of DFA across diverse experimental languages. Overall, our work presents a novel method for LLM-based MT, alleviating semantic confusion and offering promising results. Focusing on the CSUs helps LLMs to accurately re-trieve the knowledge related to the translation focus and incorporate it into the translation process, leading to the generation of high-quality translations without model training.

## Limitaion

Our work still has some limitations: 1) This study tests the impact of three types of challenging vocabulary on machine translation in the LLMs context. Additional types of challenging vocabulary can be analyzed and integrated into the method to achieve even better results. 2) The issue of semantic confusion also exists in other NLP tasks, such as dialogue systems. The proposed method can be extended in other NLP tasks to determine whether it leads to performance improvements, which might yield surprising results.

## Ethics Statement

Machine translation, stands as a pivotal task within the realm of cross-lingual NLP, boasting a broad range of applications. Our experimental approach underscores diversity by encompassing three different languages. We aim to bridge modern NLP advancements with different languages. The MT dataset utilized in our research is openly accessible. We affirm that, to the best of our knowledge, no sensitive information is disclosed, and no foreseeable risks are associated with this work.

## A Appendix

In this section, we provide some technical details.

### A.1 Languages in MT Evaluation

| Language | Code |
| --- | --- |
| English | EN |
| German | DE |
| Chinese | ZH |

Table 8: A list of languages used in our experiment and their ISO 639-1 code.

### A.2 Prompt Template

The prompt template used in this paper follows one of the most SOTA LLMs in machine translation tasks: Bayling2, as shown in Table 9.

| Baseline | |
|---|---|
| EN-ZH | 提供这句话的中文翻译： |
| ZH-EN | 提供这句话的英文翻译： |
| EN-DE | Provide the German translation for this sentence: |
| DE-EN | Provide the English translation for this sentence: |
| DFA | |
| EN-ZH | 提供这句话的中文翻译：+ semtence + [请保证下列词语的准确翻译: ] |
| ZH-EN | 提供这句话的英文翻译：+ semtence + [请保证下列词语的准确翻译: ] |
| EN-DE | Provide the German translation for this sentence: + sentence + [Ensure that the following words are accurately translated: ] |
| DE-EN | Provide the English translation for this sentence: + sentence + [Ensure that the following words are accurately translated: ] |

Table 9: Prompt template.